\documentclass{article}

\PassOptionsToPackage{compress}{natbib}

\usepackage[main, preprint]{neurips_2026}

\usepackage[utf8]{inputenc} % allow utf-8 input
\usepackage[T1]{fontenc}    % use 8-bit T1 fonts
\usepackage{hyperref}       % hyperlinks
\usepackage{url}            % simple URL typesetting
\usepackage{booktabs}       % professional-quality tables
\usepackage{amsfonts}       % blackboard math symbols
\usepackage{nicefrac}       % compact symbols for 1/2, etc.
\usepackage{microtype}      % microtypography
\usepackage{xcolor}         % colors
\usepackage{graphicx}
\usepackage{wrapfig}

\usepackage{tikz}
\usetikzlibrary{shapes.geometric, positioning, fit}
\usepackage{standalone}
\usepackage{subcaption}
\usepackage{censor}
\usepackage{enumitem}
\usepackage{algorithm}
\usepackage{algpseudocode}
\usepackage{mathtools}
\usepackage{amsmath,amssymb,amsthm}

\theoremstyle{definition}

\usepackage{listings}
\usepackage{xcolor}

\definecolor{codebg}{HTML}{F8F8F8}
\definecolor{codeframe}{HTML}{E0E0E0}
\definecolor{codecomment}{HTML}{408080}
\definecolor{codekeyword}{HTML}{008000}
\definecolor{codestring}{HTML}{BA2121}
\definecolor{codebuiltin}{HTML}{008000}
\definecolor{codenumber}{HTML}{666666}

\graphicspath{ {./figures/} }

\title{A method to automatically discover\\symbolic local learning rules}

\author{%
  Andrea Perin\thanks{Corresponding author: andrea.perin@aalto.fi. \textsuperscript{\dag}{Co-last authors.}}\\
  Aalto University\\
  \And
  Fabio Anselmi\textsuperscript{\dag}\\
  Universit\`a degli Studi di Trieste \\
  \And
  St\'ephane Deny\textsuperscript{\dag}\\
  Aalto University\\
}

\begin{document}

\newcommand{\R}{\mathbb{R}}

\maketitle

\begin{abstract}
We present a general method for discovering symbolic local learning rules that can replace backpropagation, applicable, in principle, to any deep learning architecture and task.
The method automatically discovers local loss functions specific to each layer by (1) constructing a dictionary of monomials of local tensor variables, inspired by Penrose diagram notation, (2) regressing over coefficients weighting these monomials to build layerwise polynomial losses whose gradients together minimize a target global loss.
We test the method on classification, reconstruction, and self-supervised learning tasks with small multilayer networks and minimal datasets.
Across all tasks, the discovered rules induce non-trivial interactions between layers, which lead to significant reductions in the corresponding global losses.
Furthermore, the method selects sparse combinations of monomials, finding local rules that are more interpretable than those generated by MLP-based alternatives.
The potential implications of our method range from energy-efficient deep learning to the discovery of biologically plausible learning mechanisms.
\end{abstract}

\section{Introduction}

The plausibility of backpropagation as a model of biological learning is disputed \citep[e.g.,][]{crick1989recent, lillicrap2016random, lillicrap2020backpropagation, richards2019deep}, mainly because it
demands sharing gradient information across all layers of deep networks at each update step. A more plausible candidate would have to be local and require limited information sharing across the network \citep[][]{bredenberg2023formalizing}.
From a machine learning perspective, local learning rules—when supported by suitable hardware \citep{momeni2025training}—may offer practical advantages over standard backpropagation: they reduce cross-layer information transfer, enable layers to be updated in parallel, and may reduce energy costs in hardware settings where communication is a bottleneck.

Different approaches have been proposed to replace backpropagation with local learning rules. For simple problems and shallow networks, closed-form learning rules have been derived analytically \citep[e.g.,][]{oja1982simplified,oja1989neural,hyvarinen1998independent,pehlevan2015hebbian}. 
For more complex problems and deeper networks, hand-engineered local learning rules have shown empirical success in specific settings \citep[e.g.,][]{lowe2019putting,xiong2020loco, siddiqui2023blockwise,zihan2026can}. Methods have also been proposed to \emph{automatically discover} effective local learning rules \citep[e.g.,][]{metz2018meta,gu2019meta,gregor2020,tyulmankov2022meta,metz2022velo,shervani2023meta, confavreux2023meta}. Typically, these methods train auxiliary multi-layer perceptrons (MLPs) to implement local weight updates. They, however, do not produce interpretable equations for the learning rules, as MLPs perform highly non-linear and complex transformations of their inputs.

Here, we start from the observation that many classical local learning rules can be expressed as low-order polynomial functions of locally available variables, such as weights and pre- and post-synaptic activities. 
We postulate that, for more complex problems (deeper networks and complex tasks), approximate solutions also exist that are expressible as polynomials of accessible local variables. We design a method that turns the construction of local learning rules into a systematic,  learnable procedure: for each layer, our method (1) automatically generates a dictionary of monomials of local tensor variables, leveraging a method known in physics as Penrose diagram notation, and (2) learns the coefficients weighting these elements to build polynomial local losses.
The gradients of these losses induce local updates that together conspire to satisfy a target global objective. 
We demonstrate the feasibility of our approach on minimal problems of classification, reconstruction, and self-supervised learning.

\section{Method}
We start by choosing a \emph{global loss function} $\mathcal{L}:\Theta\to\R$ that we wish to optimize, and a network architecture $f_\theta$ to train, where $\Theta$ is the space of the weights and $\theta \in \Theta$.
For each layer in the network, we define its \emph{neighborhood of degree $d$} as the set of variables including its inputs, weights, and outputs, as well as those of all the layers that are at most $d$ steps away in the downstream computational graph (Fig.~\ref{fig:fig1_schematic}).
We then construct scalar functions of elements in its neighborhood involving at most \emph{order} $K$, possibly repeated, terms.

Specifically, we consider \emph{trace monomials}, defined as those scalars that are computed by taking up to $K$, possibly repeated, neighborhood elements, and contracting all their indices two by two according to shape compatibility.
For example, for a linear layer with weight matrix $W\in\R^{d\times m}$, input $x\in\R^m$ and output $y\in\R^d$, a few possible trace monomials, of orders 2, 3, and 4, respectively, are
\begin{equation}
    ||x||^2 = \sum_i x_i x_i,
    \quad
    y^\top Wx = \sum_{ij} W_{ij}x_j y_i,
    \quad
    ||y^\top W||^2
    = \sum_{ijk} W_{ij}W_{kj}y_k y_i
    \label{eq:monomials_examples}
\end{equation}
where each contracted index appears \emph{twice and only twice} in the expression.

\begin{figure}[H]
\hfill
\begin{minipage}{0.45\textwidth}
    \begin{subfigure}[c]{\textwidth}
    	\centering
        \includegraphics{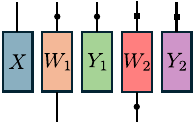}
        \caption{Examples of tensor variables in Penrose diagram notation. Vectors have a single leg, matrices have two, in general differently sized, legs.}
        \label{fig:penrose_various_boxes}
    \end{subfigure}%
\end{minipage}
\hfill
\begin{minipage}{0.45\textwidth}
    \begin{subfigure}[c]{\textwidth}
    	\centering
        \includegraphics[width=.5\textwidth]{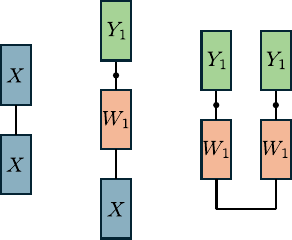}
        \caption{Trace monomials in Eq.~\ref{eq:monomials_examples} written in Penrose diagram notation.}
        \label{fig:penrose_traces}
    \end{subfigure}%
\end{minipage}
\hfill\hfill
\caption{Examples illustrating the Penrose diagram notation.}
\label{fig:penrose_example}
\end{figure}

\begin{wrapfigure}{r}{0.5\textwidth}
    \begin{center}
        \begin{tikzpicture}

    % Rectangles (weight matrices)
    \node (X) at (0,0) {$X$};
    \node[draw, rectangle, minimum width=0.5cm, minimum height=1.5cm] 
        (W1) at (1.5,0) {$W_1$};
    \node[draw, rectangle, minimum width=0.5cm, minimum height=1.5cm] 
        (W2) at (3,0) {$W_2$};
    \node[draw, rectangle, minimum width=0.5cm, minimum height=1.5cm] 
        (W3) at (4.5,0) {$W_3$};
    \node (L) at (6,0) {$\mathcal{L}$};

    % Arrows between rectangles
    \draw[->] (X) -- (W1);
    \draw[->] (W1) -- (W2);
    \draw[->] (W2) -- (W3);
    \draw[->] (W3) -- (L);
    
    \node (loss1) [draw, ellipse,
      minimum width=4cm,
      minimum height=2.5cm,
      dashed,
      color=cyan] at (1.75,0) {};
    \node [above=-.5cm of loss1, color=cyan] {$\widehat{\mathcal{L}}^1$};
    
    \node (loss2) [draw, ellipse,
      minimum width=3cm,
      minimum height=2.5cm,
      dashed,
      color=red] at (3.75,0) {};
    \node [above=-.5cm of loss2, color=red] {$\widehat{\mathcal{L}}^2$};
    
    \node (loss3) [draw, ellipse,
      minimum width=1.5cm,
      minimum height=2.5cm,
      dashed,
      color=teal] at (4.5,0) {};
    \node [above=-.5cm of loss3, color=teal] {$\widehat{\mathcal{L}}^3$};

\end{tikzpicture}
    \end{center}
    \caption{Schematic illustrating key notations used in the method. Here, the degree of the neighborhood is 2 (indicated by the ellipses).}
    \label{fig:fig1_schematic}
\end{wrapfigure}
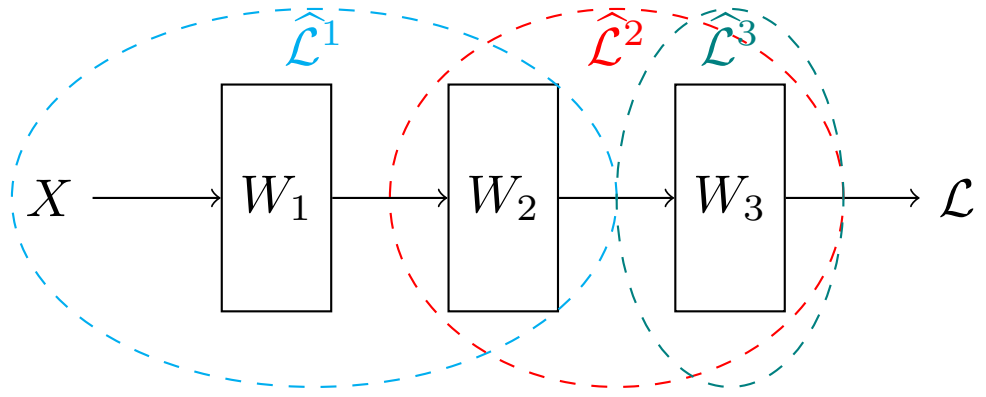

We develop a method, inspired by Penrose's tensor diagram notation \citep{penrose1971applications}, to generate all trace monomials up to order $K$. In this notation, a tensor of rank $r$ is represented as a labeled box from which $r$ different ``legs'' protrude, one for each index in the tensor, and each labeled (by equipping it with bullet points, squares, or any other graphical sign) to reflect the \emph{size} of the respective index (Fig.~\ref{fig:penrose_various_boxes}).
Two tensors can be connected by joining them on a common leg; the corresponding index is understood to be \emph{contracted}, i.e., summed over.
For a rank-0 tensor (i.e., a scalar), the box has no protruding legs.
Our method generates all combinations of up to $K$ tensors in which all legs are paired according to their shapes, leading to scalars --- the trace monomials (Fig.~\ref{fig:penrose_traces}). In all our experiments, we keep $K=4$.
In practice, we use the \texttt{einsum} function to compute the numerical values of the trace monomials (see Appendix \ref{app:tracemonomials} for a pseudo-code description of the method).

Using these monomials as dictionary elements, for each layer $\ell$ in the network we construct a \emph{local loss function}, defined as a weighted sum of its trace monomials:
\begin{equation}
\widehat{\mathcal{L}}^\ell = \sum_n \alpha_n^\ell \widehat{\mathcal{L}}^\ell_n.
\end{equation}
We refer to the weights $\alpha_n^\ell\in \R$ as \emph{rule coefficients} for layer $\ell$, and initialize them to 0. 

\setlength{\intextsep}{-1em}
\setlength{\intextsep}{0em}

We now describe a \emph{training episode}. 
We start by randomly initializing the network parameters.
Each parameter is then recursively updated by using the gradients obtained from their respective local loss function, as in standard gradient descent, for a certain number $T$ of time steps.
For each layer $\ell$ in the network (dropping the $\ell$ superscript to simplify the notation):
\begin{equation}
    \theta^{(t+1)}(\alpha) \leftarrow \theta^{(t)}(\alpha) + \sum_{n=1}^N \alpha_n \nabla_{\theta} \widehat{\mathcal{L}}_n(\alpha).
\end{equation}
We compute the global loss $\mathcal{L}$ on these resulting weights, and its gradients with respect to each layer's rule coefficients.
We update each layer's rule coefficients, and conclude the training episode.
In practice, a number $N_{\mathrm{seed}}$ of these episodes take place at once using the same rules, thus batching the process over different copies of the network parameters.

We repeat this process, each time running a new training episode with the freshly updated rule coefficients and a new set of initializations of the network parameters.
This outer loop is executed until the value of $\mathcal{L}$ is below a chosen threshold.

\subsection{Implementation details}

The meta-optimization is numerically delicate:
the outer-loop optimization is sensitive to the scale of the trace monomials, the length of the unrolled dynamics, and the stability of the induced weight updates. 
In our experiments, the following implementation choices were important for stable training.

\paragraph{Normalization of trace monomials:} Different trace monomials can have wildly different
magnitudes at initialization.
To make all of them comparable at initialization, we measure the average value of trace monomials over
$S$ initializations of the network parameters, and scale initial $\alpha$ coefficients by setting them to be the inverse of these values.

\paragraph{Increasing time horizon:} During training, we progressively increase the number of time steps $T$ in a training episode.
We also add a jitter $J$ to the number of steps within an episode for a given realization, in order to stabilize training.
We rely on \emph{checkpointed backprop} \citep{wang2009minimal,stumm2010new}, as implemented in the \texttt{Equinox} library \cite{kidger2021equinox}.
For memory reasons, this algorithm can only be run on a limited amount of data.
Thus, for each training episode, we sample a subset $B$ of the whole dataset; then, for each step in the episode, we subsample mini-batches of size $b$ from this subset.
\paragraph{Early stopping:} For a given initialization of the network parameters, a specific rule may quickly lead to divergent weights or, on the contrary, to vanishing weights.
We stop a training episode as soon as the max-norm of \emph{any} of the greedily updated weights becomes either larger or smaller than a chosen maximum/minimum safety value.
This saves compute time, but most importantly ensures that the resulting gradients for the rule coefficients are not themselves exploding or vanishing. 
For all experiments, we use a maximum of $10^2$ and a minimum of $10^{-2}$.
\paragraph{Minimax training:} At every initialization of the network parameters, we sample $N_{\mathrm{seed}}$ different copies in parallel.
    After following the current rule for the necessary number of time steps, we keep a subset $N_{\mathrm{keep}} \leq N_{\mathrm{seed}}$ of the \emph{worst-performing} seeds.
    The gradients for the rule coefficients are the average over the $N_{\mathrm{keep}}$ worst-performing seeds.
\paragraph{Optimizers:} For the inner loop, we use SGD with Nesterov momentum and unit learning rate.
    For the rule coefficients, we use AdamW with learning rate $\eta$.
\paragraph{Sparse prior on rule coefficients:} Numerically, we form rule coefficients $\alpha$ as a Hadamard product between two vectors $\alpha_1$ and $\alpha_2$ on which gradient updates are applied. This procedure results in an implicit sparse prior \citep{aladrah2026implicit} on rule coefficients, which we found to improve optimization results. As described above, $\alpha_1$ is initialized to the inverse norm of the corresponding trace monomial, and $\alpha_2$ is set to 0 at initialization.

\section{Results}

\begin{figure}
    \centering
    \includegraphics[width=\textwidth]{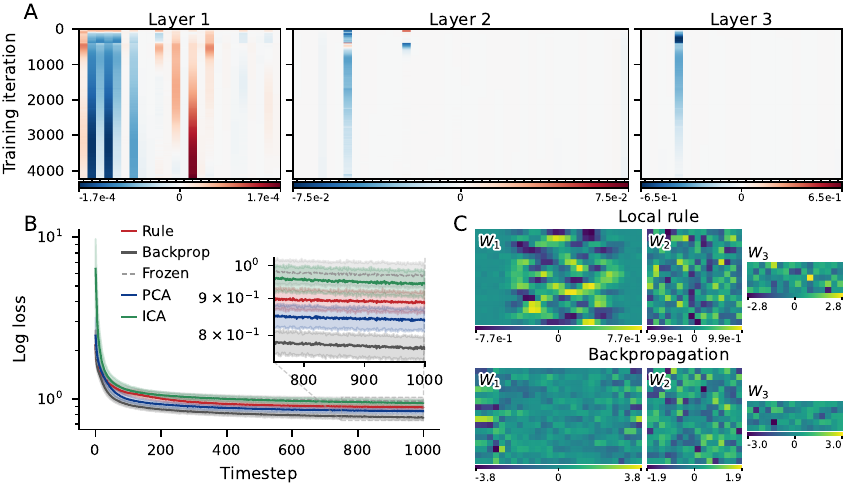}
    \caption{
    Classification task on the Sliced MNIST held-out split. (A) Evolution of $\alpha$ coefficients throughout training for a representative run. Each tick represents the $\alpha$ coefficient for one trace monomial of our dictionary. (B) Loss evolution throughout application of the learned rules (red), compared to backpropagation (grey), backpropagation with first layer frozen (dashed light grey), backpropagation with first layer frozen to the dataset's PCA (blue) and ICA (green). 
    Results are aggregated over $N_\mathrm{eval}=128$ network initializations: the median loss value is reported as a solid line, and the shaded region corresponds to the 10th–90th percentile range.    
    (C) Weights of the network after training with our learned rules (top) vs. backpropagation (bottom). Weights for PCA and ICA are shown in App. Fig.~\ref{fig:wzeros}.
    }
    \label{fig:classif}
\end{figure}

\subsection{Classification}\label{sec:class}
\paragraph{Setup} 
For classification, we consider a cross-entropy target loss with L2 regularization of weights ($\lambda_\mathrm{reg}=10^{-4}$), and use a simple, three-layer MLP with ReLU activations in the first and second layers, and log-softmax activation in the last layer.
Thus, the classification portion of the loss function is the negative dot product between the network's output and the one-hot encoded label.

To keep computational load minimal, we used a low-dimensional version of MNIST, \textit{Sliced MNIST}\label{slicedMNIST}, in which we extract the middle (14th) row of pixels in the image, resulting in inputs of dimension 28.
In order to make the classification problem more challenging, we randomly flip the sign of each sample vector: this last step makes the problem unsolvable by linear methods. We standardize the dataset to 0 mean and unit variance before using it.
We then split the dataset into a training split consisting of classes (0,1,2,3,4), and a held-out split consisting of classes (5,6,7,8,9).
We apply the same pipeline to Fashion-MNIST, of which we keep classes (5,6,7,8,9); the rules we find on the training split are then used both on the held-out split and on the Sliced Fashion-MNIST dataset, to test how the rules we find \textit{transfer} across similar datasets.
We choose the following weight sizes: $W_1: (16, 28), W_2: (16, 16), W_3: (5, 16)$.
We use $N_\mathrm{seed}=512$, $N_\mathrm{keep}=64$, $\eta=3\times 10^{-5}$, $B=2048, b=256$, $J=0$, and $T$ ranging from 1 to 500 over 4000 training episodes, and then staying at 500 for an additional 200 training episodes. 

We consider neighborhoods of all possible degrees: for degree 1, the rules are \textit{purely local}, as there are no cross-layer interactions. Conversely, at degree 3, all layers have access to the whole downstream computational graph, thus rules are \textit{global}.
At degree 2, the rules include nearest-neighbor interactions: layers 2 and 3 have access to the whole downstream computational graph, while layer 1 is limited to itself and layer 2. Thus, degree 2 is \textit{local}, but not purely.
By sweeping over the possible degrees, we show how the method can interpolate between purely local and global rules.

\paragraph{Results}
We test the learned rules by using them to train $N_\mathrm{eval}=128$ different copies of the same architecture we trained on, using the held-out splits of Sliced MNIST and Sliced Fashion-MNIST.
The results for degree 2 on the held-out split of Sliced MNIST are reported in Fig.~\ref{fig:classif}.
Results for the other data splits, including Sliced Fashion-MNIST, are shown in App. Fig.~\ref{fig:class_streams}.
To meaningfully evaluate our method, we compare its performance to four alternatives:
(1) standard backpropagation, (2) keeping the first layer frozen to its initialization value, and training only the last two layers (hereby called \textit{frozen backprop}), (3) keeping the first layer frozen to the dataset's principal components (PCA)---a learner implementing a generalized Hebbian/Sanger rule will converge to this solution \citep{SANGER1989459, oja1989neural}, and (4)  keeping the first layer frozen to the independent components (ICA)---a learner implementing a version of Oja's rule will converge to this solution \cite{hyvarinen1998independent}.
Among these baselines, frozen backpropagation constitutes a lower threshold target: at degree 2, the last two layers have access to the target loss, which can be written as a dot product between the one-hot encoded labels and the net's outputs.
Thus, we expect that a local update rule for layer 1 will \textit{improve} on the results achievable by frozen backprop.
The remaining baselines, instead, serve to provide a comparison to other local learning methods.

When local rules learned by our method are applied, the loss decreases more slowly than it does with backpropagation, but faster than frozen backpropagation---thus achieving the minimal result (see Fig.~\ref{fig:classif}B).
Our rule also outperforms the ICA baseline, but not the PCA baseline (rule: 0.89, frozen: 0.97,  backprop: 0.77, PCA: 0.84, ICA: 0.94).
In terms of accuracy, our rule achieves 67\% (backprop: 71\%, frozen: 63\%, PCA: 68\%, ICA: 65\%).
These results are remarkable, since classification tasks are known to be especially challenging to solve with local learning rules (see \cite{Baldi2015A} and Discussion).

A feature of our method is that it affords explicit, symbolic expressions for the local losses.
In the following, we denote by $\hat{Y}$ the one-hot encoded labels.
We restrict ourselves to a few of the top-magnitude terms for clarity\footnote{In the following, a norm operator applied to a matrix is understood to be a Frobenius norm.}:
\begin{equation}
\begin{aligned}
\widehat{\mathcal{L}}^1
 = 10^{-5}\big(&
  16.9\,\lVert W_1^\top Y_1 \rVert^2
  - 15.6\,Y_1 \cdot (Y_1 + W_1 X)
  - 8.94\,\lVert W_1 X \rVert^2 \\
  &- 1.48\,Y_2 \cdot (Y_2 + W_2 Y_1)
  + 1.37\,\lVert W_1 \rVert^2
  - 1.20\,\lVert X \rVert^2 \lVert Y_1 \rVert^2
\big) \\
\widehat{\mathcal{L}}^2
 = 10^{-3}\big(&
  -9.31\,\Hat{Y}^\top W_3 Y_2
  - 7.90\,\Hat{Y}\cdot Y_3
  + 1.04\,(\Hat{Y}\cdot Y_3)^2
\big) \\
\widehat{\mathcal{L}}^3
 = 10^{-2}\big(&
  -2.50\,\Hat{Y}^\top W_3 Y_2
  - 1.88\,\Hat{Y}\cdot Y_3
\big)
\end{aligned}
\end{equation}
Layers 2 and 3 are able to pick up the target loss (the term $\Hat{Y}\cdot Y_3$), as well as a logit alignment monomial ($\Hat{Y}^\top W_3 Y_2$).
In the case of layer 2, we also recognize a squared cross-entropy term, whose gradient is aligned with the cross-entropy but with inverse sign.
Layer 1 does not have access to the target loss terms, since the degree is only 2.
In its loss, instead, we recognize terms that promote output magnitude ($Y_1 \cdot (Y_1 + W_1 X)$), together with others which moderate the dynamics ($\lVert W_1^\top Y_1 \rVert^2$ and $\lVert W_1 \rVert^2$, for instance),
and forward-looking terms which contain layer 2's quantities, such as $Y_2 \cdot (Y_2 + W_2 Y_1)$.
The latter terms are effectively a way for layer 1 to adapt to the features learned by layer 2, and thus may represent a surrogate of the usual backprop's backward channel.

\subsection{Reconstruction}

\begin{figure}
    \centering
    \includegraphics[width=\textwidth]{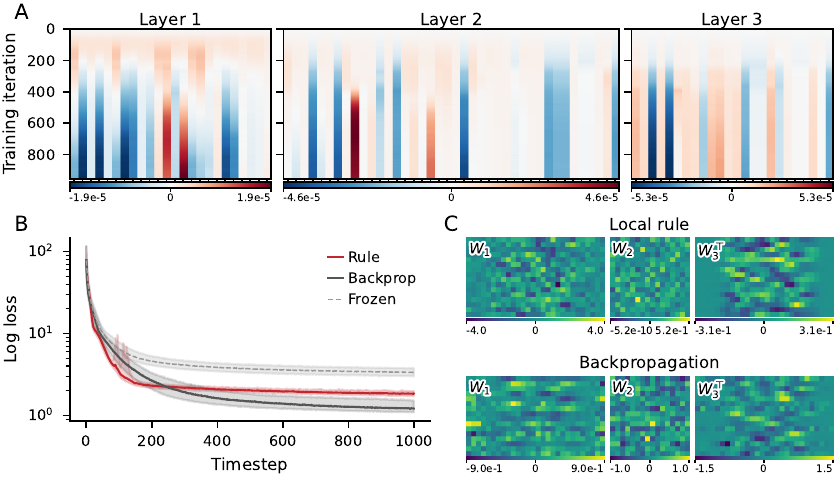}
    % \caption{
    % Reconstruction task on the Sliced MNIST held-out split. (A) Evolution of $\alpha$ coefficients throughout training for a representative run. Each tick represents the alpha coefficient for one trace monomial of our dictionary. 
    % (B) Loss evolution throughout application of the learned rules (red), compared to backpropagation (grey) and backpropagation with first layer frozen (dashed light grey).
    % Results are aggregated over $N_\mathrm{eval}=128$ network initializations: the median loss value is reported as a solid line, and the shaded region corresponds to the 10th-90th quantile range.    
    % (C) Weights of the network after training with our learned rules (top) vs. backpropagation (bottom).
    % }
    \caption{Reconstruction task. Same caption as Fig.~\ref{fig:classif}.}
    \label{fig:reconstruct}
\end{figure}

\paragraph{Setup} We train a non-linear autoencoder on a reconstruction task on Sliced MNIST (see \ref{slicedMNIST}).
We use a 3-layer MLP with sizes $W_1: (16, 28), W_2: (16, 16), W_3: (28,16)$, ReLU activations on layers 1 and 2, and no nonlinearity on the last layer.
The target is the L2 loss between samples and their reconstructions.

As in the classification case, we train the rules on classes (0,1,2,3,4) of Sliced MNIST, and test them on the held-out classes (5,6,7,8,9), and on Sliced Fashion-MNIST.
We use $N_\mathrm{seed}=512, N_\mathrm{keep}=256, B=2048, b=256, \eta=10^{-5}$, $J=5$, and $T$ ranging from 1 to 300 over 2000 training episodes.

\paragraph{Results} During training, $\alpha$ coefficients are selected by the method (Fig.~\ref{fig:reconstruct}A), leading to interpretable local loss equations.
As local rules are subsequently applied, the loss decreases more slowly than it does with backpropagation.
However, it decreases faster than it does with a degree-2 local version of backpropagation, where the first layer is frozen to initial weights and only the two last layers are trained with backpropagation (Fig.~\ref{fig:reconstruct}B).
The fact that our method beats this frozen backprop setup demonstrates that the learned losses cooperate in a non-trivial way across all layers, beyond what backpropagation limited to a neighborhood of degree-2 can accomplish (average loss values are, for rule: 1.90, frozen: 3.35, backprop: 1.27).
The learned weights are structurally similar to those obtained with backpropagation (Fig.~\ref{fig:reconstruct}C).
Results for the other data splits, including Sliced Fashion-MNIST, are shown in App. Fig.~\ref{fig:rec_streams}.

We report the largest terms in the rules (the cutoff being chosen for legibility):
\begin{equation}
\begin{aligned}
\widehat{\mathcal{L}}^1
 = 10^{-5}\big(&
-1.85 \lVert W_1 X \rVert^2 
-1.83  Y_1 \cdot (Y_1 + W_1X)
+1.81  \lVert Y_1^\top W_1 \rVert^2\\
&-1.65 \lVert W_2\rVert^2 \lVert Y_1 \rVert^2
-1.31 \lVert Y_1 \rVert^2 \lVert X \rVert^2
-1.20  \lVert W_2 Y_1 \rVert^2
+1.20  \lVert W_1^\top W_1\rVert^2
\big) \\
\widehat{\mathcal{L}}^2
 = 10^{-5}\big(&
  -4.30  (W_3^\top X)\cdot (W_3^\top Y_3)
  +4.18 \lVert W_2 Y_1\rVert^2 
  -4.07 \lVert W_2\rVert^2 X\cdot Y_3
  -3.78 X\cdot (Y_3 + W_3 Y_2)\\
  &+2.55 \lVert Y_2^\top W_2\rVert^2
  -2.50 \lVert X\rVert^2 X\cdot Y_3 
  -2.47  (W_3^\top X)\cdot (W_2 Y_1)
  -2.45 \lVert W_3\rVert^2 X\cdot Y_3
\big) \\
\widehat{\mathcal{L}}^3
 = 10^{-5}\big(&
  -5.23 X \cdot (W_3 Y_2 + Y_3)
  -2.82 \lVert W_3\rVert^2 X\cdot Y_3
  -2.73 \lVert X\rVert^2 X\cdot Y_3
\big)
\end{aligned}
\end{equation}
In this case, the target loss is not directly available as a dictionary element to \emph{any} layer, being the squared difference $\lVert X-Y_3\rVert^2$.
However, under assumptions of fixed norms $\lVert X\rVert^2$ and $\lVert Y_3 \rVert^2$, it becomes proportional to $-X \cdot Y_3$.
This latter term is found in layers 2 and 3, as well as terms which regularize the norm of $Y_3$.
In layer 1, which does not have access to the final output $Y_3$, there are Hebbian-like terms (such as $\lVert W_1 X \rVert^2 $ and $\lVert Y_1\rVert^2 $), reconstruction-promoting terms ($Y_1\cdot W_1 X$ and $\lVert Y_1^\top W_1 \rVert^2$), and forward-looking terms which contain $W_2$, such as $\lVert W_2 Y_1 \rVert^2$ and $\lVert W_2\rVert^2 \lVert Y_1 \rVert^2$. The latter terms are effectively a way for layer 1 to adapt to the features learned by layer 2, and thus may represent a surrogate of the usual backprop's backward channel.

\subsection{Self-supervised learning}

\begin{figure}
    \centering
    \includegraphics[width=\textwidth]{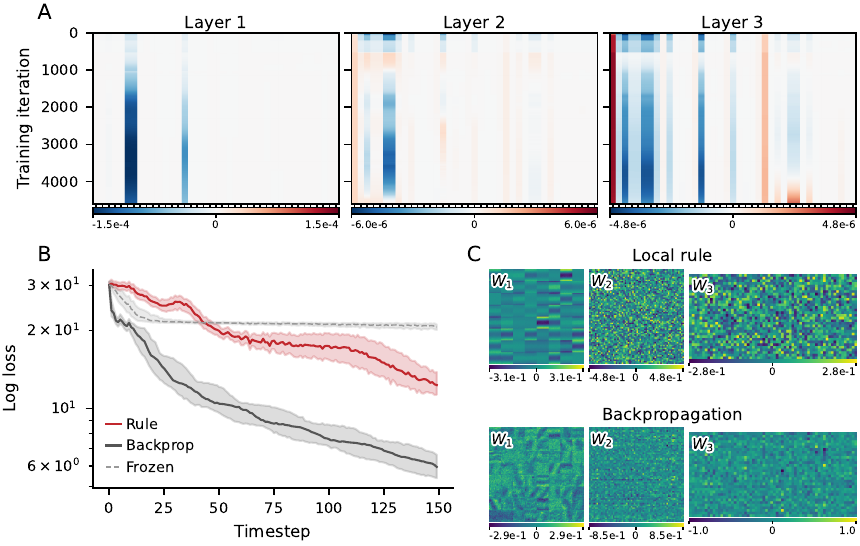}
    \caption{Self-supervised task on the MNIST held-out split. 
    (A) Evolution of $\alpha$ coefficients throughout training for a representative run. Each tick represents the $\alpha$ coefficient for one trace monomial of our dictionary. 
    (B) Loss evolution throughout application of the learned rules (red), compared to backpropagation (grey) and backpropagation with \emph{first two layers} frozen (dashed light grey).
    Results are aggregated over $N_\mathrm{eval}=128$ network initializations: the median loss value is reported as a solid line, and the shaded region corresponds to the 10th–90th percentile range.
    We evaluate the rules for approximately the number of time steps they were trained for.
    (C) Weights of the network after training with our learned rules (top) vs. backpropagation (bottom). $W_1$ has been rearranged to show the weights of each neuron as $28\times 28$ patches.
    }
    \label{fig:ssl}
\end{figure}

\paragraph{Setup}
We train a 3-layer MLP on a self-supervised learning task on full MNIST, flattening the digits into 784-dimensional input vectors.
We consider random flips and shifts of up to 157 pixels (corresponding to a fifth of the total input length) as augmentations.
The MLP has sizes $W_1: (64, 784), W_2: (64, 64), W_3: (32, 64)$, and we use the Barlow Twins loss function \citep{zbontar2021barlow} with tradeoff parameter $\lambda = 0.0625$, and with L2 regularization of weights ($\lambda_\mathrm{reg}=10^{-3}$).
We use $N_\mathrm{seed}=128, N_\mathrm{keep}=128, B=1024, b=256, \eta=5\times 10^{-6}$, $J=5$, and $T$ ranging from 1 to 100 over 6000 training episodes. We keep $T$ to a smaller value than in previous experiments due to stability issues in this setup.

For this task, we choose to employ \emph{layerwise local} learning rules, i.e., we choose a neighborhood of degree 1. Indeed, recent work \citep{zihan2026can} suggests that backpropagation could be approximated with layerwise local learning rules for typical self-supervised learning objectives.
\paragraph{Results}
Local learning rules are learned on a training set of classes (0,1,2,3,4), and subsequently applied to a separate testing set of classes (5,6,7,8,9). Results on this held-out set are reported in Fig.~\ref{fig:ssl}. As before, when local rules learned by our method are applied, the loss decreases more slowly than it does with backpropagation, but faster than it does with a degree-1 local version of backpropagation where the first two layers are frozen to initial weights (average loss values are, for rule: 17.11, frozen: 22.47, backprop: 5.50).
Results for the other data splits, including Fashion-MNIST, are shown in App. Fig.~\ref{fig:ssl_streams}.

We report the largest terms in the rules (the cutoff being chosen for legibility).
Terms with an apex denote one of the two augmentation pairs; we note how the method finds very similar coefficients for terms that are swap-invariant.
\begin{equation}
\begin{aligned}
\widehat{\mathcal{L}}^1
 = 10^{-5}\big(&
 -12.3 (Y_1'\cdot (W_1 X) + Y_1 \cdot (W_1 X'))
 -6.16 (W_1 X)\cdot (W_1 X')
 +1.56 \lVert W_1 \rVert^2
\big) \\
\widehat{\mathcal{L}}^2
 = 10^{-7}\big(&
 12.4 \lVert W_2\rVert^2
 +9.22 (Y_2\cdot Y_2')
 +5.94 (Y_2 \cdot (W_2 Y_1 + Y_2))
 +5.92 (Y_2' \cdot (W_2 Y_1' + Y_2'))\\
 &+4.60 Y_2\cdot (W_2 Y_1')
 +4.59 Y_2'\cdot (W_2 Y_1)
\big) \\
\widehat{\mathcal{L}}^3
 = 10^{-6}\big(&
 +5.72 (W_3^\top Y_3)\cdot(W_3^\top Y_3')
 +4.00 \lVert W_3\rVert^2\\
 &-2.82 (Y_3\cdot Y_3' + Y_3'\cdot(W_3 Y_2) + Y_3\cdot(W_3 Y_2') + (W_3 Y_2)\cdot (W_3 Y_2'))
\big)
\end{aligned}
\end{equation}
In these formulae, we recognize terms linked to the cross-augmentation pre- and post-nonlinearity (e.g., $Y_1'\cdot (W_1X)$ and $Y_1\cdot(W_1X')$, note the symmetry between augmented pairs) and others that are linked to activation magnitudes (e.g., $||W_2Y_1||^2$).
Terms are also present that measure the cross-augmentation alignment, e.g., $(W_1 X)\cdot (W_1 X')$.
We also note that many rule coefficients are matched in value.
For these coefficients, we also observed that they typically evolve in a grouped manner during the optimization process, showing the same values throughout training. We take this ``coefficient grouping'' as a sign that our method is well-behaved. 
Indeed, in the self-supervised learning framework we use, inputs and outputs are always present in transformation pairs.
To the model, there is no inherent difference between samples in an augmentation pair, as they share the same shape and statistics.
Note that in layer 3 the grouped monomials are strictly equivalent due to the absence of nonlinearity in the final layer of our network.

\subsection{Effects of degree and order}
\begin{figure}[t]
    \centering
    \begin{subfigure}[t]{0.48\textwidth}
        \centering
        \includegraphics[width=\linewidth]{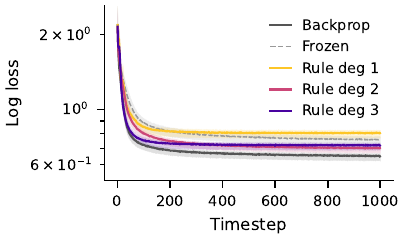}
        \label{fig:class_comp}
    \end{subfigure}
    \hfill
    \begin{subfigure}[t]{0.48\textwidth}
        \centering
        \includegraphics[width=\linewidth]{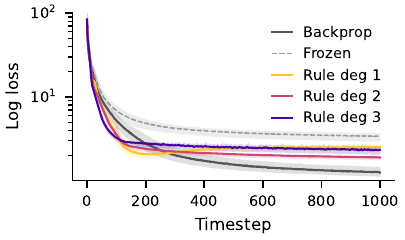}
        \label{fig:rec_comp}
    \end{subfigure}
    \caption{Effect of degree on the rules' performance as evaluated on the same-split of Sliced MNIST for classification (left) and reconstruction (right).}
    \label{fig:degree}
\end{figure}

For both classification and reconstruction tasks, we perform \textit{degree sweeps}: holding other hyperparameters fixed, we vary the degree, ranging from 1 (purely local rules) to 3 (global rules) (Fig.~\ref{fig:degree}).
Interestingly, we find that degree 2 can comfortably beat degree 1 on the classification task, but not on the reconstruction task, where degree 1 is competitive.
This seems to corroborate previous theoretical work \cite{Baldi2015A} on the necessity of a backward communication channel for solving supervised tasks.
Moreover, in both cases rules of degree 2 outperform global rules, but not full backpropagation.
This shows that our global rules are not able to exactly recover backpropagation terms, even when they are explicitly available.
Future work on the optimization side of the method could potentially address this limitation.

Another important hyperparameter in our method is the \textit{order} of the monomials, which we fixed to $K=4$ in all our experiments.
Increasing the order results in larger dictionaries, and increases the expressive power of a ruleset (at the cost of increased computational burden).
In App. Table~\ref{tab:cnts} we provide the number of monomials as a function of degree and order for each of the layers in the MLP described in Sec.~\ref{sec:class}.
We leave a detailed exploration of the effects of this hyperparameter for future work.

\subsection{Shape invariance}
The construction of the dictionary is shape-invariant: the number and the nature of the monomials do not depend on the specific shapes of the involved tensors.
A natural question then is: are rules also shape-invariant?
To test this, we run the degree-2 rules for the classification task on the held-out split of Sliced MNIST, on a network with different layer sizes than the network used for learning the rule (described in \ref{sec:class}), specifically, $W_1: (32, 28), W_2: (16, 32), W_3: (5, 16)$. We report the results in App. Fig.~\ref{fig:classif_shape}.
The rules transfer partially to this differently sized architecture, but the performance deteriorates and stability seems to worsen.
We leave further investigations on shape-invariance for future work.

\section{Discussion}
We presented a general method to learn equations for local learning rules. 
Other symbolic regression methods, such as SINDy,  share a similar aim of inferring equations from data~\citep{DeepSR2019,AIFeynamn_2_2020,Discovering_GNNs2020,Neural_Transformers2021,Transformers_2_2022,GP_SR_review2023,Review_scientific_discovery2022,LLM_SR2025,doi:10.1073/pnas.1517384113}.
They are usually employed to reverse-engineer dynamical systems from observations of their trajectories.
Therein lies a crucial distinction: our method does not use trajectory data as a ground truth signal, and instead \emph{relies solely on the signal given by the global loss}.
A trajectory-based method, on the contrary, would pose both practical as well as theoretical problems: practical because of the data requirements (a whole trajectory, in our case, would consist of many copies of the weights of a model), and theoretical because what we want to approximate is not the backpropagation-induced dynamics, but instead the backpropagation-induced \emph{performance}. In analogy with physics, we wish to apply constraints on the \emph{macroscopic observable of the complex system} (the network's performance as measured by the global loss) rather than on its microscopic configurations.

Most previous attempts at meta-learning local rules have used MLPs \citep[e.g.,][]{metz2018meta,gu2019meta,gregor2020,tyulmankov2022meta,metz2022velo,shervani2023meta,confavreux2023meta}, trained to learn the gradients associated with specific layers in a pre-specified network architecture.
Our method offers advantages over MLP-based ones. First, once the target architecture is specified, we find expressions which are more \emph{interpretable} (as demonstrated in the Results section) than previous methods. 
Second, our method offers promises for transferring rules learned on small architectures and low-dimensional datasets to larger architectures and higher-dimensional datasets. Once written symbolically, the rules we find for the gradients carry no explicit dependence on the tensor shapes of data or network. Our preliminary results indicate that rules learned on one architecture can, at least partially, transfer to another architecture similar to the one used for training. Finally, our design choice of using a dictionary of low-order monomials in the local variables also distinguishes our method from the few other existing symbolic meta-learning methods, which restrict themselves either to variants of the Hebbian rule \citep[e.g.,][]{lindsey2020learning,sandler2021meta}, or to the estimation of optimal hyperparameters for normal backpropagation \citep[e.g.,][]{zheng2022symbolic}.

Our attempts at finding polynomial local learning rules should not be interpreted as assuming that such rules are sufficient for arbitrary deep supervised learning, as limitations of local learning are known \citep{Baldi2015A}.
In our framework, locality is a tunable parameter rather than a hard constraint: increasing the neighborhood degree gives each layer access to progressively larger portions of the downstream computational graph, up to the global case where the target loss itself can be represented at each layer.
Indeed, if the target loss can be expressed as a linear combination of the trace monomials, our method can potentially recover backpropagation as one possible learning rule, where each layerwise loss is the target loss itself.
Our framework is thus not limited to using purely local rules, but can be used to quantify how much nonlocal information is needed to obtain useful learning dynamics. 
Indeed, we have started exploring the effects of tuning the degree of our rules, and found corroborations to existing theoretical predictions.
In addition to locality, our method allows us to tune the order of the trace monomials.
Higher-order trace monomials can produce more complex local gradients, at the cost of increased computational burden and less stable dynamics.

Future work will also focus on scaling up the method to more realistic architectures and datasets.
Improving the stability of the training process should be prioritized, as it is currently quite sensitive to hyperparameter choices, in part because the optimization landscape is fractal \cite{sohl2024boundary}.
Another promising direction is the analysis of the algebraic structure and interpretability of the rules learned by the method and, crucially, their \emph{transferability} across datasets and architectures. If such transferability holds, the method could be used to discover local rules on simple datasets and then apply them to more complex ones. Lastly, we mention a possible alternative training method, known as the \emph{adjoint} method \citep[as used in, e.g.,][]{bai2019deep}, which relies on the implicit function theorem, and unlike the greedy procedure we presented, does not require to keep track of the whole trajectory.
Instead, it requires reaching a fixed point of the dynamics, after which the gradients can be obtained by solving a linear system.
\vfill

\section*{Acknowledgments}
We thank Jaakko Lehtinen for useful discussions, and the Aalto Science-IT project for providing the computational resources. 
Funding: Research Council of Finland grant 3357590 (“Neuroscience-inspired Deep Learning”) to S.D.
The authors declare no competing interests.

\medskip

\bibliographystyle{abbrvnat}
\bibliography{references}

%%%%%%%%%%%%%%%%%%%%%%%%%%%%%%%%%%%%%%%%%%%%%%%%%%%%%%%%%%%%

\appendix

\section{Method to list trace monomials}
\label{app:tracemonomials}
The following code snippet shows the way we enumerate trace monomials.
\begin{lstlisting}[caption={Function to generate trace monomials.}, label={lst:tracegen}]
from collections import Counter
from typing import Tuple, Generator, Sequence, List
from sympy import Symbol
import itertools as its


# Type aliases
Shape = Tuple[Symbol, ...]
T = Tuple[str, Shape]
Signature = Tuple[Tuple[str, ...], Tuple[Tuple[int, ...], ...]]


def pairings(things: List[int]):
  """
  >>> list(pairings([1, 2, 3, 4]))
  [[(1, 2), (3, 4)], [(1, 3), (2, 4)], [(1, 4), (2, 3)]]
  """
  if len(things) == 2:
    yield [tuple(things)]
  else:
    for i in range(1, len(things)):
      yield from (
        [(things[0], things[i])] + p
        for p in pairings(things[1:i] + things[i + 1 :])
      )


def generate_traces(tlist: Sequence[T], n: int):
  assert n > 1
  combos = its.combinations_with_replacement(tlist, n)
  for tensors in combos:
    splitpoints = (0,) + tuple(its.accumulate([len(s) for n, s in tensors]))
    tnames = tuple(n for n, s in tensors)
    all_idxs = tuple(its.chain.from_iterable(t[1] for t in tensors))
    counts = Counter(all_idxs)
    if all(v % 2 == 0 for k, v in counts.items()):
      positions = [
        [p for p, sym in enumerate(all_idxs) if sym == s] for s in counts
      ]
      allpairings = its.product(*tuple(map(pairings, positions)))
      for p in allpairings:
        assignments = [-1 for t in all_idxs]
        p = its.chain.from_iterable(p)
        for v, (p1, p2) in enumerate(p):
          assignments[p1] = assignments[p2] = v
        assignments = [
          assignments[splitpoints[idx] : splitpoints[idx + 1]]
          for idx in range(len(tnames))
        ]
        assignments = tuple(tuple(a) for a in assignments)
        is_invalid = any(
          Counter(a).most_common(1)[0][1] > 1 for a in assignments
        )
        if not is_invalid:
          yield (tnames, assignments)
\end{lstlisting}

\section{Trace monomial counts}
\begin{table}[th]
  \centering
  \caption{Monomial counts in a 3-layer MLP as a function of degree and order.}
  \begin{tabular}{c ccc}
    \toprule
    Order & Degree 1 & Degree 2 & Degree 3 \\
    \midrule
    2 & $[2,2,3]$     & $[3,4,3]$      & $[5,4,3]$      \\
    3 & $[3,3,5]$     & $[5,7,5]$      & $[9,7,5]$      \\
    4 & $[11,11,24]$  & $[24,40,24]$   & $[58,40,24]$   \\
    5 & $[18,18,42]$  & $[45,78,42]$   & $[120,78,42]$  \\
    6 & $[57,57,155]$ & $[171,328,155]$& $[536,328,155]$\\
    \bottomrule
  \end{tabular}
  \label{tab:cnts}
\end{table}

\section{Additional results}

\begin{figure}[th]
    \centering
    \includegraphics[]{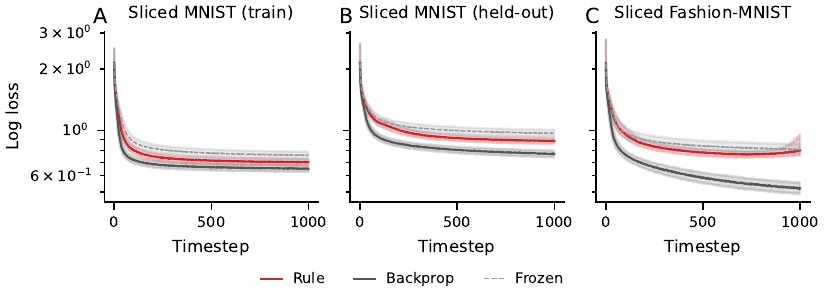}
    \caption{Comparison of the performance of the learned degree-2 rules for the classification task on (A) the same Sliced MNIST split that was used for training, (B) the held-out split of Sliced MNIST, and (C) Sliced Fashion-MNIST.
    Performance is best on the same and held-out splits, while it degrades slightly at the tail end of the training on the Fashion-MNIST split.}
    \label{fig:class_streams}
\end{figure}

\begin{figure}[ht]
    \centering
    \includegraphics[]{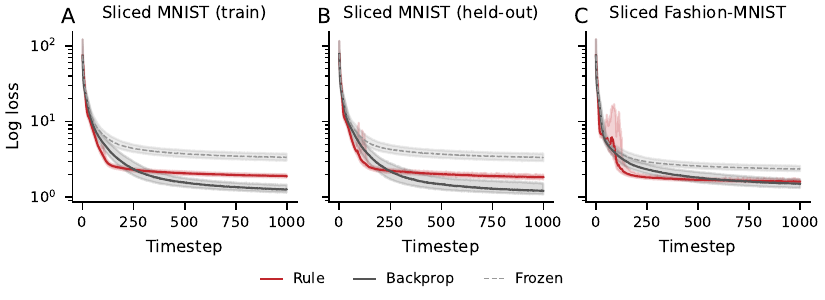}
    \caption{Comparison of the performance of the learned degree-2 rules for the reconstruction task on (A) the same Sliced MNIST split that was used for training, (B) the held-out split of Sliced MNIST, and (C) Sliced Fashion-MNIST.
    Performance is similar across tasks, with slightly higher instability on the Sliced Fashion-MNIST split.}
    \label{fig:rec_streams}
\end{figure}

\begin{figure}[ht]
    \centering
    \includegraphics[]{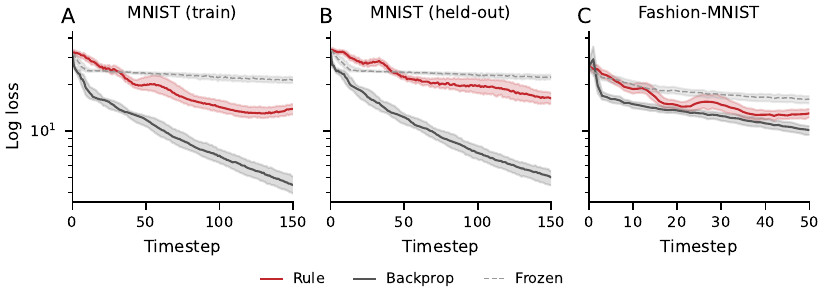}
    \caption{Comparison of the performance of the learned degree 1 rules for the self-supervised learning task on (A) the same MNIST split that was used for training, (B) the held-out split of MNIST, and (C) Fashion-MNIST.
    We stopped training early on the Fashion-MNIST split due to stability issues.
}
    \label{fig:ssl_streams}
\end{figure}

\begin{figure}
    \centering
    \includegraphics{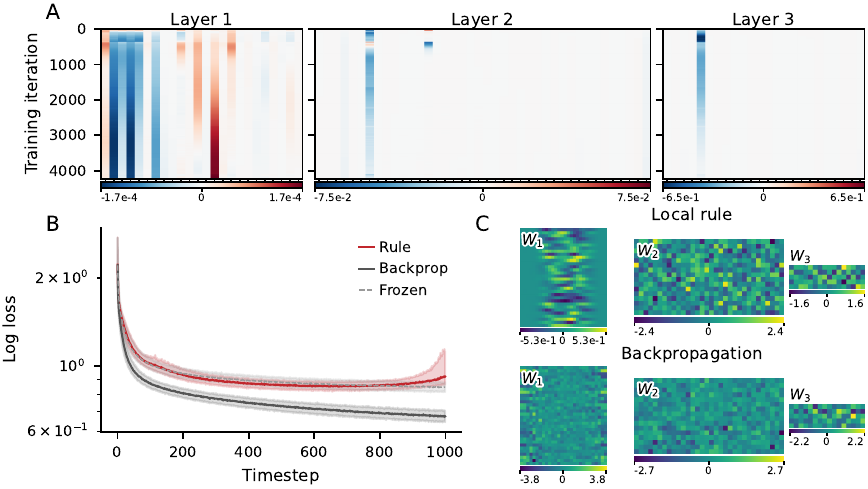}
    \caption{Testing the degree-2 classification rules on a differently sized network.}
    \label{fig:classif_shape}
\end{figure}

\begin{figure}[t]
    \centering
    \begin{subfigure}[t]{0.48\textwidth}
        \centering
        \includegraphics[width=\linewidth]{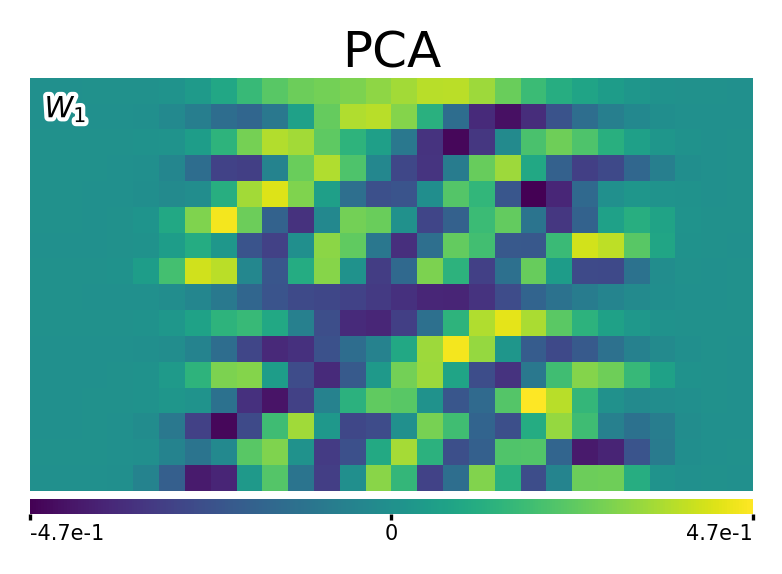}
    \end{subfigure}
    \hfill
    \begin{subfigure}[t]{0.48\textwidth}
        \centering
        \includegraphics[width=\linewidth]{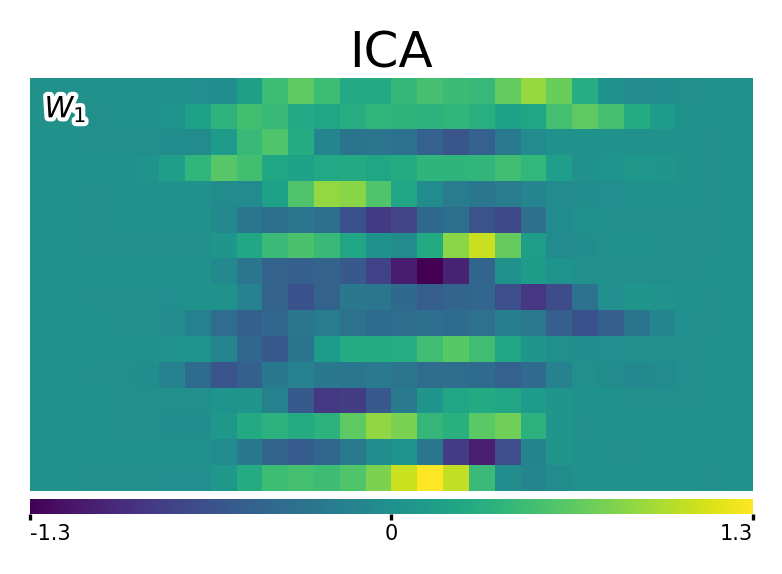}
    \end{subfigure}
    \caption{First frozen layer for the PCA (left) and ICA (right) cases in the held-out split of Sliced MNIST.
    In order to compose the first layer, we first run PCA/ICA with a number of components that is half the layer's output dimension; we then stack them
    with a copy multiplied by $-1$. We do so to address the presence of the ReLU nonlinearity.
    }
    \label{fig:wzeros}
\end{figure}

\end{document}